\documentclass[10pt,conference]{IEEEtran}
\IEEEoverridecommandlockouts

\usepackage{comment}

\usepackage[numbers,sort&compress]{natbib}
\usepackage{amsmath,amssymb,amsfonts}
\usepackage{graphicx}
\usepackage{textcomp}
\usepackage[table]{xcolor}
\usepackage{subcaption}
\usepackage[hidelinks]{hyperref}
\usepackage{balance}
\usepackage{cleveref}
\usepackage{dblfloatfix}
\usepackage[font={small,it}]{caption}
\usepackage{tabularx}
\usepackage{float}
\usepackage{svg}
\usepackage{makecell}
\usepackage{enumitem}
\usepackage{pifont}
\usepackage{multirow}
\usepackage{booktabs}
\usepackage{arydshln}
\usepackage{colortbl}
\usepackage{diagbox}
\usepackage{longtable}
\usepackage[ruled,vlined]{algorithm2e}
\usepackage[cache=false]{minted}
\usepackage{microtype}
\definecolor{codebg}{RGB}{245,245,245}
\setminted{
  linenos,
  breaklines,
  breakanywhere,
  autogobble,
  fontsize=\footnotesize,
  bgcolor=codebg,
  frame=leftline,
  framesep=1.5mm,
  numbersep=8pt,           
  xleftmargin=2.6em,
  tabsize=4
}

\def\BibTeX{{\rm B\kern-.05em{\sc i\kern-.025em b}\kern-.08em
    T\kern-.1667em\lower.7ex\hbox{E}\kern-.125emX}}

\makeatletter

\newcommand*\titleheader[1]{\gdef\@titleheader{#1}}
\AtBeginDocument{%
  \let\st@red@title\@title
  \def\@title{%
    \bgroup\normalfont\large\centering\@titleheader\par\egroup
    \vskip1.5em\st@red@title}
}

\makeatother

\makeatletter
 \let\old@ps@headings\ps@headings
 \let\old@ps@IEEEtitlepagestyle\ps@IEEEtitlepagestyle
\def\confheader#1{%
  \def\ps@headings{%
    \old@ps@headings%
  }%
  \def\ps@IEEEtitlepagestyle{%
    \old@ps@IEEEtitlepagestyle%
    \def\@oddhead{%
      \begin{tabular}{@{}l@{}}
        \fontsize{8pt}{10pt}\selectfont\strut #1 \\ % Added \fontsize here
        \strut \hfill
      \end{tabular}%
    }%
    \def\@evenhead{%
      \begin{tabular}{@{}l@{}}
        \fontsize{8pt}{10pt}\selectfont\strut #1 \\ % Added \fontsize here
        \strut \hfill
      \end{tabular}%
    }%
  }%
  \ps@headings%
}

\makeatother

\begin{document}
\title{On the Impact of Feature Reduction Techniques on Adversarial Attacks}

\author{
\IEEEauthorblockN{Mohammad Meymani, Roozbeh Razavi-Far}
\IEEEauthorblockA{
Trustworthy and Secure AI (TSAI) Lab\\
Faculty of Computer Science, University of New Brunswick, Fredericton, Canada\\
\{mohammad.meymani79, roozbeh.razavi-far\}@unb.ca
}
}
\maketitle

\begin{abstract}
Adversarial attacks have posed significant threats to machine learning models by exploiting their inherent vulnerabilities. Researchers have established an interdisciplinary field called adversarial machine learning to explore these vulnerabilities and develop corresponding defense strategies. Feature reduction strategies are used to decrease the dimensionality of the data in order to reduce training time, avoid the curse of dimensionality, prevent overfitting, and mitigate redundant features. However, their effect on adversarial robustness across different classifiers and attack scenarios remains insufficiently explored. This study aims to analyze the effect of feature reduction techniques on the adversarial robustness of a diverse set of machine learning models under poisoning, backdoor, and universal adversarial perturbation attacks on the APT dataset. These models include support vector machine, decision tree, random forest, multi-layer perceptron, and k-nearest neighbors. The attained results are compared with scenarios where different feature reduction techniques are applied, including principal component analysis, permutation importance ranking, recursive feature elimination, and univariate mutual information. Metrics such as accuracy, F1-score, and mitigation score are employed to evaluate the effect of these techniques on adversarial robustness. The results demonstrate that permutation importance ranking and univariate mutual information generally provide the most consistent robustness improvements, while other methods exhibit more variable effects depending on the classifier and attack scenario.
\end{abstract}
\section{Introduction}
{M}achine {l}earning (ML) is a branch of {a}rtificial {i}ntelligence (AI), which has revolutionized many domains including security, {n}atural {l}anguage {p}rocessing (NLP), retrieval, autonomous driving, healthcare, {c}omputer {v}ision (CV), finance, and many other domains. While ML models have facilitated many tasks, their privacy and utility are threatened by adversarial attacks. An adversarial attack can possess different amounts of knowledge about the target model. An adversary with a perfect knowledge is considered as a white-box adversary, while having no knowledge is considered as a black-box adversary. Moreover, an adversary with partial knowledge is called a gray-box adversary. {A}dversarial {m}achine {l}earning (AML) studies the vulnerabilities of ML models, adversarial attacks, and their corresponding countermeasures \cite{qiao2026survey,meymani2026defense,machado2021adversarial,meymani2026divided}. 

Adversarial attacks, in a general view, are divided into evasion, exploratory, or poisoning attacks. Evasion attacks aim to degrade the performance of ML models during inference, by adding subtle and carefully crafted noise to the test data. Exploratory attacks violate the privacy of the ML models. They aim to probe the ML models and steal their sensitive information. Finally, poisoning attacks employ different strategies by altering the features and/or labels of the training data during training. The goal of poisoning attacks is to stay stealthy during the training while significantly degrading the ML model's performance during the inference \cite{qiao2026survey,meymani2026defense,razavi2026quantum,hallaji2026fednia}. Moreover, backdoor poisoning attacks aim to embed malicious backdoor triggers to the training data in order to degrade the predictive task performance during inference, where target patterns or labels are present \cite{qiao2026survey,meymani2026defense,vashagh2026recent}. 

{F}eature {s}election (FS) in ML refers to the techniques, which select the most informative features. FS aims to improve ML models' accuracy, reduce overfitting, decrease training time by removing irrelevant and redundant features. {D}imensionality {r}eduction (DR) techniques reduce the dimensionality of the data by creating a new representation while maintaining as much variability as possible from the original data. Both FS and DR methods help with avoiding the curse of dimensionality and facilitate visualization. There are numerous FS/DR techniques, which analyze the data from different perspectives. Some techniques aim to preserve a certain amount of the original data variance, some others set a predefined number of features, and a few techniques take classification labels into consideration \cite{theng2024feature,jia2022feature,van2009dimensionality}.

{Although FS/DR methods are widely used to increase the efficiency and generalization capabilities of ML models, their effect on adversarial robustness remains unexplored. In particular, it is unclear whether feature reduction mitigates adversarial effects or introduces new vulnerabilities across different classifiers and attack scenarios. Existing studies mainly focus on adversarial feature selection in specific settings, while a systematic comparison of conventional FS/DR techniques across multiple classifiers and attack types remains limited \cite{ozfatura2021less,zhang2015adversarial,zhang2024improving}. This study addresses this gap by evaluating the effects of four FS/DR techniques on the adversarial robustness of five ML classifiers under poisoning, backdoor, and universal adversarial perturbation (UAP) attacks. Our contributions are as follows:}

\begin{itemize}
    \item We evaluate the robustness of five different classifiers under different attack scenarios.
    \item We analyze the behavior of the classifiers in adversarial environments, where feature reduction is present.
    \item We introduce a new metric called \textit{mitigation score}, which evaluates the robustness gain after applying FS/DR techniques.
\end{itemize}

{The rest of this paper is organized as follows. Section \ref{sec:related-works} reviews the related research studies. Section \ref{sec:problem-settings} discusses the background and problem settings. Section \ref{sec:experimental-setups} describes the experimental methodology and evaluation settings. Section \ref{sec:discussion-and-results} presents and discusses the attained results. Section \ref{sec:conclusion} concludes the paper. }
\section{Related Work}
\label{sec:related-works}
%In \cite{bao2018mitigation}, the authors use Lasso ($\ell_1$ regularization), which is an embedded feature selection, to assess the adversarial robustness of a logistic regression classifier. The authors initially suggest that reducing the feature space can result in shrinking the adversarial region available for the attacker and decreasing the model's complexity. The authors use {J}acobian-based {S}aliency {M}ap {A}ttack (JSMA) to generate adversarial examples in $\ell_1$, $\ell_2$, and $\ell_\infty$ norms. Moreover, the authors compute {M}aximum {M}ean {D}iscrepancy (MMD) between clean and adversarial samples, where larger MMD makes adversarial example detection easier. The authors employ MNIST, Ransomware, and PDF Malware as the datasets. The results indicate that a modest reduction in accuracy can substantially increase the robustness. 

In \cite{ozfatura2021less}, the authors argue that only a subset of latent features are consistent under adversarial perturbation; as a result, the authors aim to identify robust latent features in penultimate layers rather than relying on the input features. The authors then propose {l}atent {m}asking (LM) strategy, which calculates the difference in each neuron activation value before and after launching adversarial attack. The neurons with the least changes are considered robust. Experiments on Resnet-18 on CIFAR-10 and CIFAR-100 datasets demonstrate substantial robustness gains against {p}rojected {g}radient {d}escent (PGD) attacks.

In \cite{zhang2015adversarial}, the authors propose {w}rapper-based {a}dversarial {f}eature {s}election (WAFS) method in order to protect against evasion attacks. WAFS formulates feature selection as multi-objective optimization problem, which maximizes both generalization capability and robustness. The authors then employ {h}arness {o}f {e}vasion (HoS) as the security metric, which calculates the smallest perturbation that can change the classification result. So, the security is ultimately calculated as the average of the HoS over the samples. Experimental results on both spam filtering and PDF malware detection datasets demonstrate that WAFS provides more robustness compared to the traditional wrapper-based feature selection methods.

In \cite{zhang2024improving}, the authors propose {a}dversarially {r}obust {e}nsemble {c}lassifier (AREC), which includes three main stages: diversified adversarial feature selection, training a pool of robust classifiers, and random aggregation during inference. The first stage selects features one-by-one to preserve the trade-off between accuracy and robustness. AREC stores a feature vector bank, where at every iteration all the selected features till that iteration are stored as a feature subset. As a result, the first feature vector contains a single best feature, the second contains the top two features, and the final vector contains all the best features. Every feature vector from the bank is used to train a {s}upport {v}ector {m}achine (SVM) classifier. During inference, a subset of SVM classifiers are chosen for every distinct prediction. The results indicate the significant mitigation of white- and black-box evasion attacks across spam filtering, malware detection, and handwritten digit recognition datasets.

{Although these studies demonstrate the potential of adversarially informed feature selection, they mainly focus on developing specialized methods for improving robustness against evasion attacks. In contrast, this study evaluates how conventional feature reduction techniques affect adversarial robustness across multiple classifiers under different attack scenarios.}
\section{Problem Settings}
\label{sec:problem-settings}
{P}rinciple {C}omponent {A}nalysis (PCA) is an unsupervised dimensionality reduction technique that transforms a set of features into a smaller set called principal components. Principal components aim to preserve as much of the original data variance as possible by computing eigenvalues and eigenvectors. Eigen vectors demonstrate the orthogonal directions of maximum variance and the corresponding eigenvalues show the amount of the captured variance. The principal components are ranked from highest to the lowest of the captured variance, allowing us to select the most informative principal components \cite{salem2019data,abdi2010principal}.

{P}ermutation {i}mportance {r}anking (PIR) is a model-agnostic feature selection method, which computes the importance of each feature by randomly permuting the values of that feature to observe the degradation of the model's performance. The decrease in the target metric (accuracy or F1-score) demonstrates the importance of a feature; a substantial performance drop suggests a more important feature, while little or no change indicates an unimportant feature. Although PIR does not rely on model-specific parameters and assumptions, its performance can degrade when the features are highly correlated \cite{altmann2010permutation,zubair2024new}.

{R}ecursive {f}eature {e}limination (RFE) is a wrapper-based feature selection method that recursively removes the least important features according to the importance scores provided by a trained ML model. The process begins by training a model using all available features and ranking them based on a model-specific importance measure. The least important feature (or a predefined number of features) is then eliminated, and the model is retrained on the remaining feature subset. This iterative procedure continues until the desired number of features is reached, producing a ranked list of features based on their contribution to the performance of the model. RFE often achieves a compact and highly informative feature subset that can improve the model's interpretability and reduce overfitting. However, since the model needs to be retrained at every iteration, RFE can be computationally expensive for high-dimensional datasets \cite{li2005analysis,priyatno2024systematic}.

{U}nivariate {m}utual {i}nformation (UMI) is a filter-based feature selection method that evaluates the relevance of each feature independently by measuring its {m}utual {i}nformation (MI) with the final output. Mutual information measures the amount of information that a feature provides about the target by obtaining both linear and nonlinear statistical dependencies. A higher mutual information score indicates that knowing the value of the feature reduces the uncertainty about the target, while a score close to zero suggests little or no predictive relevance. In UMI, features are ranked according to their individual mutual information scores, and the top-ranked features are then selected for downstream model training. Because each feature is evaluated independently, UMI is computationally efficient, model-agnostic, and suitable for high-dimensional datasets \cite{michel2008mutual,doquire2013mutual}.

SVM is a supervised ML algorithm used for classification and regression tasks, especially for high-dimensional datasets. The goal of the SVM is to find an optimal decision boundary, which is called maximum-margin hyperplane. That hyperplane separates different classes by maximizing the overall distance (margin) between the hyperplane and the nearest samples (support vectors). SVM originally solves linear problems, however, by employing kernel functions, it maps data to higher dimensional feature space in order to linearly separate them in that latent representation \cite{hearst1998support,awad2015support}.

{D}ecision {t}ree (DT) is a supervised ML algorithm, which performs classification and regression by iteratively splitting the dataset based on features. DT employs a tree-like structure, which includes decision nodes, branches, and leaf nodes. Decision nodes split data based on feature values, branches represent the outcome of data splitting, and leaf nodes show the final predicted class for classification or a numerical value for regression \cite{safavian1991survey}.

\begin{figure*}[h]
    \centering
    \includegraphics[width=0.9\linewidth]{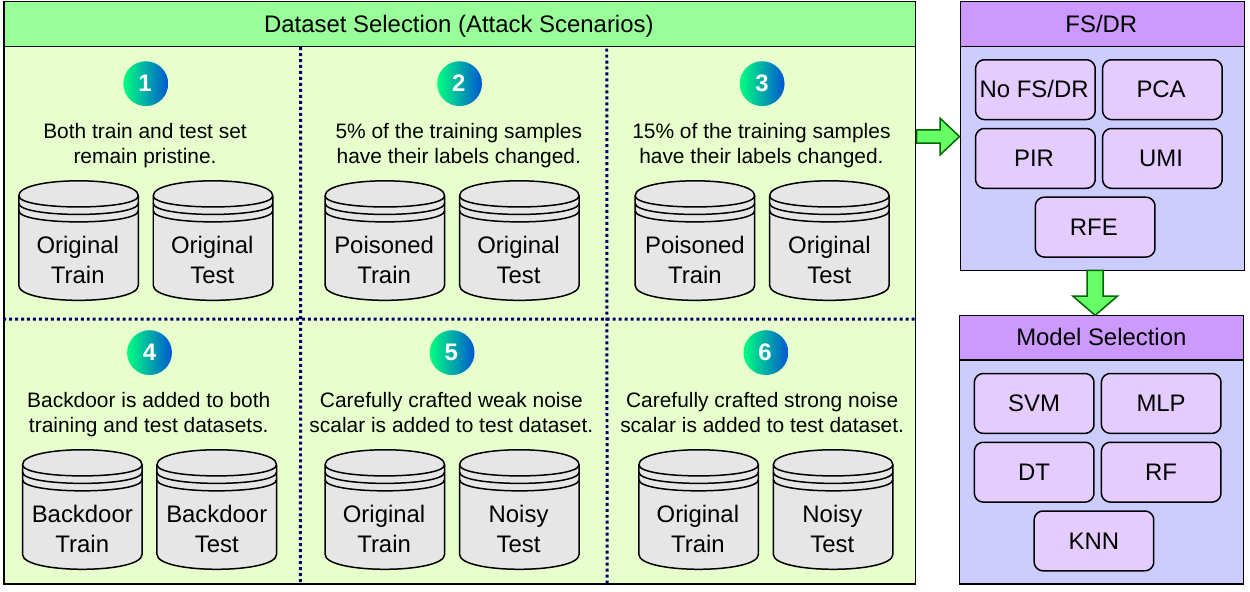}
    \caption{Simulated scenarios and experimental methodology.}
    \label{fig:experimental-methodology}
\end{figure*}

{R}andom {f}orest (RF) is an ensemble learning algorithm that includes DTs as base learners, and combines their predictions to make the final decision. During training, each DT is built using bootstrap sampling, where at every split a random subset of features is chosen to increase the diversity among the DTs. For classification tasks, the final decision is made via majority or weighted voting \cite{breiman2001random,biau2016random}.

{M}ulti-{l}ayer {p}erceptron (MLP) is a feedforward artificial neural network that consists of an input layer, one or more hidden layers, and an output layer. MLPs can learn hierarchical feature representations from data, reducing the need for manual feature tuning. However, they usually need a larger training dataset, greater computational resources, and careful hyperparameter tuning to achieve optimal performance \cite{xiang2005geometrical}.

{K}-{n}earest {n}eighbors (KNN) is an instance-based supervised learning algorithm used for classification and regression tasks. Unlike model-based methods, KNN does not build a predictive model during training; instead, it stores the training data and predicts based on the similarity between a test sample and the training samples. For classification, the algorithm identifies the $k$ nearest neighbors of a test sample according to a distance metric, and assigns the class that receives the majority vote among the neighbors \cite{guo2003knn,halder2024enhancing}.

{This study employs PCA, PIR, RFE, and UMI for feature reduction and SVM, DT, RF, MLP, and KNN for classification. These settings ensure a varied set of classifiers and feature reduction techniques in order to analyze the results properly.}  
\section{Methodology and Experimental Setups}
\label{sec:experimental-setups}
This section describes the experimental methodology, datasets, FS/DR methods, classifiers, selected hyperparameters, and evaluation metrics. Figure \ref{fig:experimental-methodology} illustrates the step-by-step process of conducting experiments, where each classifier has been trained on each dataset using 10-fold cross-validation process. After that, the best hyperparameters are chosen and each classifier is trained on the entire training data, and is evaluated on the final test data.
\subsection{Attacks}
Three white-box attack types are considered: label flipping poisoning, backdoor, and UAP. In all attack scenarios, the adversary has perfect knowledge about the training/test data and the target model.

Two corruption rates of $5\%$ and $15\%$ are considered for label flipping. Samples near the SVM decision boundary are randomly selected, and their labels are flipped. In the targeted backdoor poisoning attack, a trigger value of 0.01 is added to the odd-indexed features of 20\% of the training data, and their labels are changed to the target label. For the test data, the trigger was added to all of the samples without affecting the label values. 

{In UAP attacks, the universal scalar perturbation is optimized using a grid-search procedure over 1000 uniformly spaced perturbation values within a predefined interval $[-\epsilon,\epsilon]$. For each value, the average log-loss is computed across the ensemble of classifiers, and the perturbation that maximizes this objective is selected. The optimal perturbation scalar is then added to the test set.  Two perturbation budgets (0.25 and 0.5) are used to evaluate the models' performance against both strong and weak perturbations.}
\subsection{Simulated Scenarios}
We use APT \cite{laurenza2017malware} dataset, which includes 16 class labels, more than 260 features, and more than 11,000 samples; then, we split the data into $80\%$ train and $20\%$ as the final test. Missing values are imputed using the mean strategy, and the data are standardized across all scenarios using a standard scaler (mean = 0, standard deviation = 1). The experiments follow six scenarios. In the first scenario, both train and test datasets remain unchanged to evaluate the performance of the models and FS/DR techniques in a non-adversarial environment. The second and the third scenarios assess the performance of training data poisoning under $5\%$ and $15\%$ corruption rates.

Scenario four tests the performance of hidden backdoor attack. We evaluate the performance, where backdoor is present in both train and test datasets. Scenarios five and six demonstrate the performance, where carefully crafted adversarial noise scalar is added to the test dataset. These noise values are generated during the training session and are added to the final test dataset.
\subsection{Model and Hyperparameter Selection}
PCA, PIR, RFE, and UMI are employed as FS/DR methods, while SVM, DT, RF, MLP, and KNN are used as classifiers. For each classifier and experimental scenario, 10-fold cross-validation is applied to identify the best hyperparameters during training. Each model is then trained from scratch on the entire training dataset using the selected hyperparameters and evaluated on the test dataset. The same approach is employed for FS/DR methods as well. {To prevent data leakage, all preprocessing and FS/DR techniques are fitted exclusively on the training data within each cross-validation fold. After hyperparameter selection, these techniques are refitted on the entire training set, and the same fitted transformations are applied to the test set without using its samples or labels during fitting.}
\subsection{Metrics}
Three evaluation metrics are employed: accuracy, macro-F1, and mitigation score.
\subsubsection{Accuracy}
It is calculated by dividing the number of correct predictions by the total number of predictions:
\begin{equation}
    \text{Accuracy} = \frac{\# \text{The Correct Predictions}}{\# \text{The Total Number of Predictions}}.
\end{equation}
\subsubsection{F1-score}
It measures the performance of a classifier by calculating the harmonic mean of precision and recall metrics. The value for F1-score sits between zero and one, where a higher value suggests a better ML model.

Precision for class $c$ is defined as the fraction of correctly classified samples of class $c$ among all samples predicted as class $c$:
\begin{equation}
    \text{Precision}_c = \frac{\text{TP}_c}{\text{TP}_c + \text{FP}_c},
\end{equation}
where TP denotes true positive and FP denotes false positive.

Recall for class $c$ is defined as the fraction of correctly classified samples of class $c$ among all samples that have actual label of $c$:
\begin{equation}
    \text{Recall}_c = \frac{\text{TP}_c}{\text{TP}_c + \text{FN}_c},
\end{equation}
where FN denotes false negative.

F1-score presents a combination of Precision and Recall to provide a more stable metric, and for each class is calculated as follows:
\begin{equation}
    \text{F1-score}_c = \frac{2 \times \text{Precision}_c \times \text{Recall}_c}{\text{Precision}_c+\text{Recall}_c}.
\end{equation}

Macro-F1 calculates the average of F1-score values for each class in order to demonstrate the final performance of the classifier:
\begin{equation}
    \text{Macro-F1} = \frac{1}{|C|} \sum_{c \in C} \text{F1-score}_c,
\end{equation}
where $C$ is a set including all the classes.
\subsubsection{Mitigation Score}
It measures how an adversarial attack is mitigated when a FS or DR algorithm is applied on data. By having scores as the input values, mitigation score is calculated as follows: 
\begin{equation}
    \text{Mitigation Score} = (\mathcal{S}_\text{Clean}^\text{Before} - \mathcal{S}_\text{Attack}^\text{Before}) - (\mathcal{S}_\text{Clean}^\text{After} - \mathcal{S}_\text{Attack}^\text{After}),
\end{equation}
where $\mathcal{S}_\text{Clean}^\text{Before}$ is the score on the clean data before FS/DR, $\mathcal{S}_\text{Attack}^\text{Before}$ is the score on the adversarial data before FS/DR, $\mathcal{S}_\text{Clean}^\text{After}$ is the score on the clean data after FS/DR, and $\mathcal{S}_\text{Attack}^\text{After}$ is the score on the adversarial data after FS/DR. 

{A positive mitigation score indicates that FS/DR reduces the performance degradation caused by an attack, while a negative score indicates increased degradation. A score of zero indicates no change in attack-induced degradation. Therefore, the mitigation score measures the relative robustness improvement rather than the absolute classification performance.}
\section{Results and Discussion}
\label{sec:discussion-and-results}
This section presents the experimental results and analyzes the behavior of the feature reduction techniques for each classifier under the simulated attack scenarios.

\begin{figure}[h]
    \centering
    \includegraphics[width=0.9\linewidth]{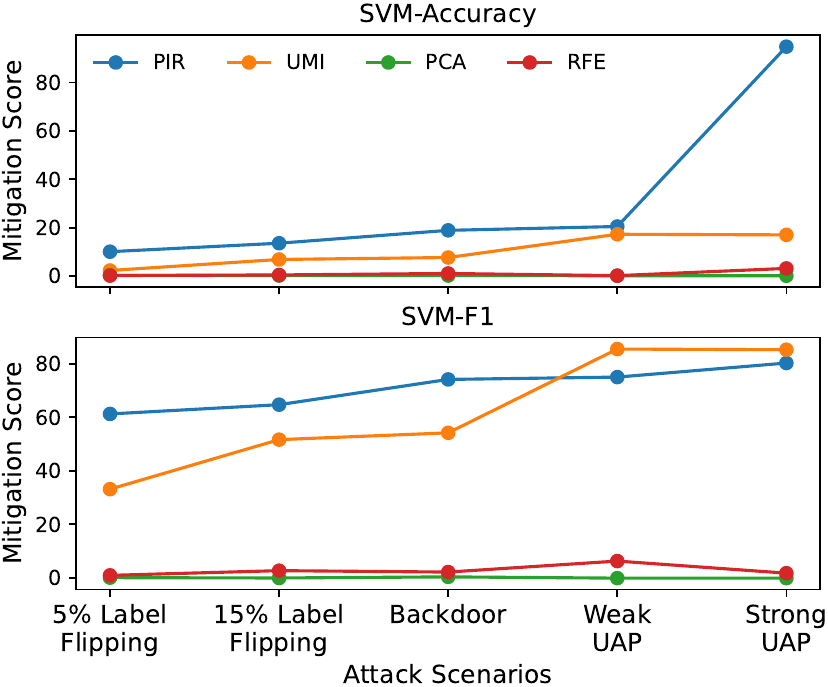}
    \caption{Mitigation scores of SVM in terms of accuracy and Macro-F1.}
    \label{fig:svm}
\end{figure}

\textbf{SVM Results}: Figure \ref{fig:svm} illustrates the mitigation scores of the SVM classifier in terms of both accuracy and Macro-F1 across the different attack scenarios. For the accuracy, all four methods achieve positive mitigation scores across every scenario, indicating that FS/DR methods consistently reduce the impact of adversarial attacks on the SVM classifier. Among the evaluated techniques, PIR provides the largest and most consistent robustness improvements, particularly against the UAP attacks, while UMI also demonstrates substantial mitigation scores across most attack scenarios. In contrast, RFE and PCA provide only limited improvements, suggesting that eliminating or transforming features without explicitly considering their predictive importance is less effective for improving adversarial robustness. Macro-F1 follows almost the same trend as accuracy, except for PCA, which achieves insignificant negative mitigation scores for $15\%$ label flipping and UAP attacks.

\begin{figure}[h]
    \centering
    \includegraphics[width=0.9\linewidth]{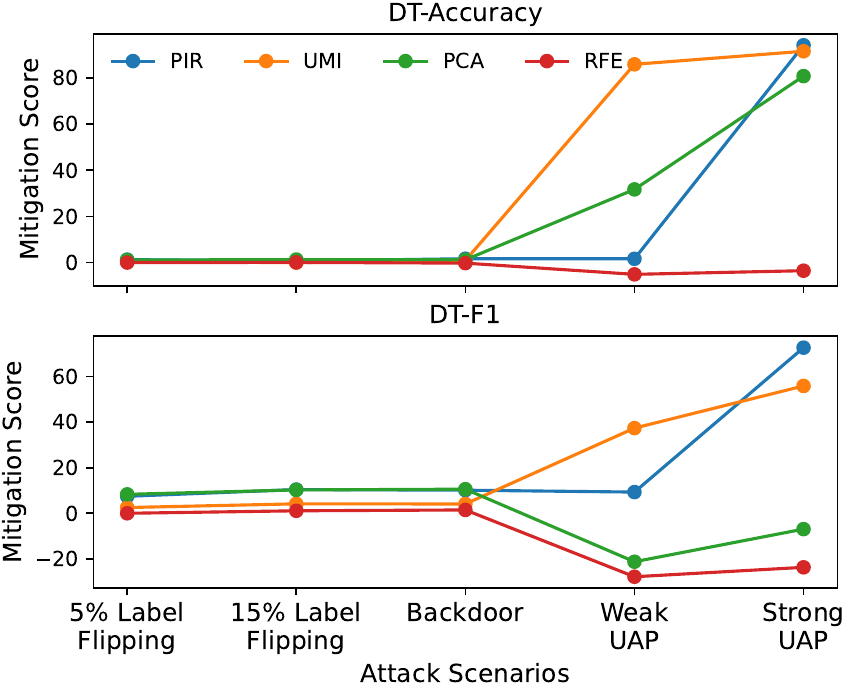}
    \caption{Mitigation scores of DT in terms of accuracy and Macro-F1.}
    \label{fig:DT}
\end{figure}

\textbf{DT Results}: Figure \ref{fig:DT} presents the mitigation scores of the DT classifier in terms of both accuracy and Macro-F1 across the different attack scenarios. For accuracy, FS/DR methods exhibit distinct behaviors depending on the attack type. Under label flipping and backdoor attacks, all methods produce only marginal mitigation scores (mostly positive), indicating that feature reduction has a limited influence on the robustness of DT against poisoning-based attacks. In contrast, all methods except RFE exhibit positive mitigation scores under UAP attacks, demonstrating that feature reduction is considerably more effective against perturbation-based attacks than poisoning attacks for DT. Among the evaluated methods, UMI consistently achieves the strongest improvements under weak UAP attacks, while PIR becomes the most effective approach under strong UAP attacks. PCA also provides substantial mitigation against UAP attacks {in terms of accuracy}, whereas RFE produces negative mitigation scores, indicating that RFE further increases the degradation caused by adversarial perturbations. The Macro-F1 results reveal a more varied behavior across the evaluated methods. PIR and UMI consistently maintain positive mitigation scores across all attack scenarios, demonstrating stable improvements in preserving balanced classification performance. PCA and RFE demonstrate positive mitigation scores on poisoning-based attacks, while having a negative effect on UAP attacks.

\begin{figure}[h]
    \centering
    \includegraphics[width=0.9\linewidth]{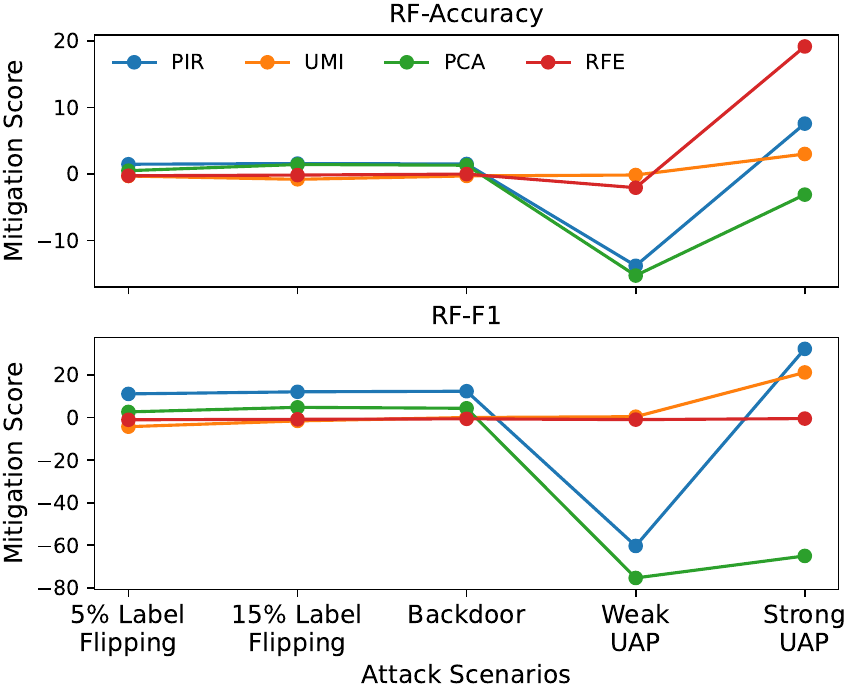}
    \caption{Mitigation scores of RF in terms of accuracy and Macro-F1.}
    \label{fig:RF}
\end{figure}

\textbf{RF Results}: Figure \ref{fig:RF} presents the mitigation scores of the RF classifier in terms of both accuracy and Macro-F1 across the evaluated attack scenarios. For accuracy, none of the feature selection or dimensionality reduction methods consistently improve robustness across all attacks. PIR provides stable positive mitigation against label flipping and backdoor attacks and achieves additional improvement under strong UAP attack, but its performance degrades under weak UAP attack. UMI exhibits negative mitigation scores for all scenarios except strong UAP attack. PCA shows negligible positive mitigation scores for poisoning-based attacks, and negative scores for UAP attacks. RFE yields negative mitigation scores across most attack settings. Overall, the accuracy results indicate that the robustness gains of RF are highly attack-dependent and considerably less consistent than those observed for SVM and DT. Macro-F1 almost shows the same trend as accuracy.

\begin{figure}[h]
    \centering
    \includegraphics[width=0.9\linewidth]{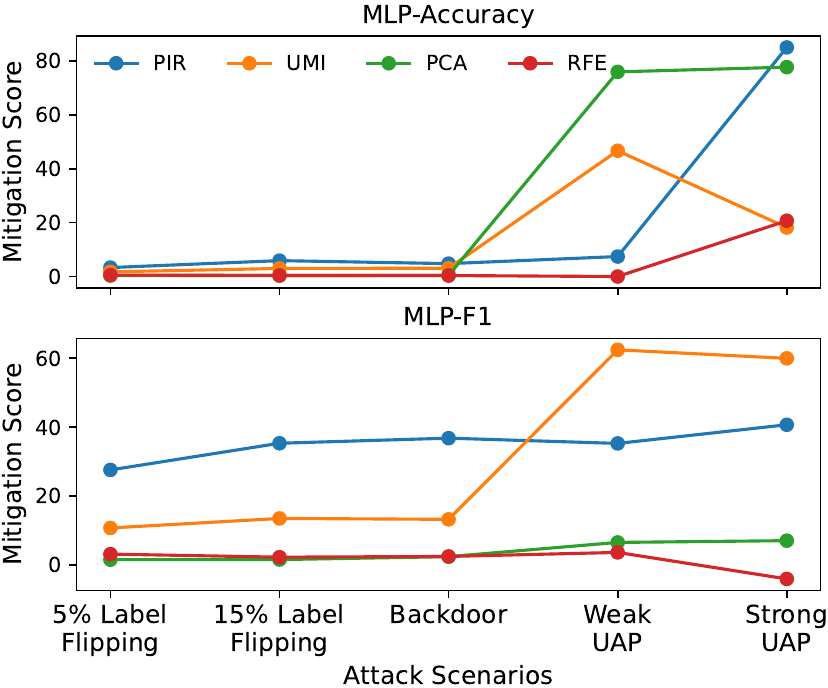}
    \caption{Mitigation scores of MLP in terms of accuracy and Macro-F1.}
    \label{fig:MLP}
\end{figure}

\textbf{MLP Results}: Figure \ref{fig:MLP} presents the mitigation scores of the MLP classifier in terms of both accuracy and Macro-F1 across the evaluated attack scenarios. For accuracy, all four feature reduction methods provide moderate positive mitigation scores under poisoning-based attacks, and substantial positive mitigation scores under UAP attacks. However, Macro-F1 tells a different story, where only PIR and UMI gain substantial positive mitigation scores under all attack scenarios. PCA achieves moderate robustness gains under all attack scenarios. RFE achieves positive mitigation scores under all attack scenarios, except for strong UAP attack.

\begin{figure}[h]
    \centering
    \includegraphics[width=0.9\linewidth]{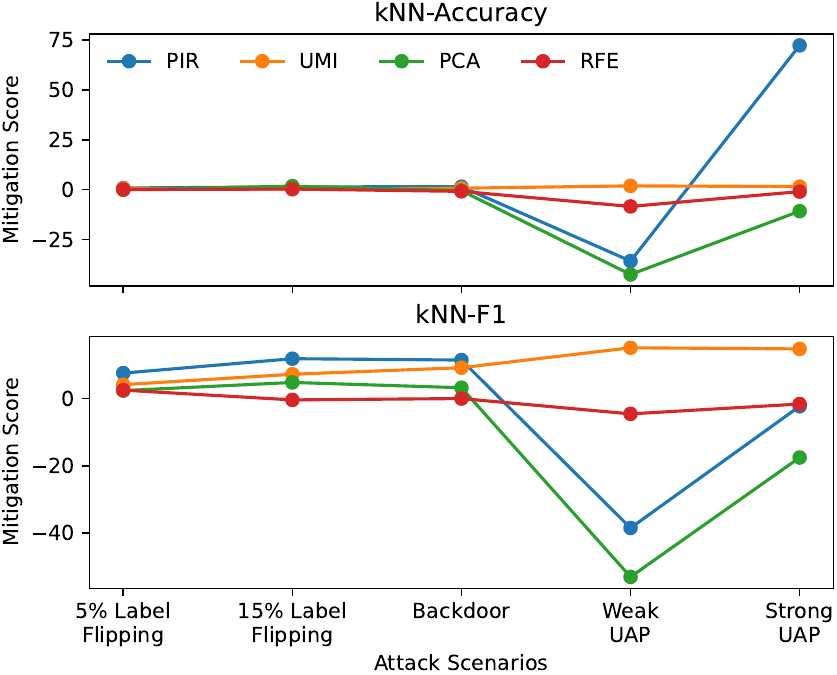}
    \caption{Mitigation scores of KNN in terms of accuracy and Macro-F1.}
    \label{fig:KNN}
\end{figure}

\textbf{KNN Results}: Figure \ref{fig:KNN} presents the mitigation scores of the KNN classifier in terms of both accuracy and Macro-F1 under the different adversarial attack scenarios. For accuracy, all feature reduction methods provide only marginal mitigation against label flipping and backdoor attacks, indicating that reducing the feature space has little influence on the robustness of KNN against poisoning-based attacks. A substantially different behavior is observed under UAP attacks. PIR achieves the largest improvement under strong UAP attacks, whereas UMI maintains positive mitigation under both weak and strong UAP attacks. In contrast, PCA produces large negative mitigation scores under both UAP attack scenarios. RFE degrades robustness under weak UAP attack. For Macro-F1, UMI achieves positive mitigation scores across all scenarios. All other methods achieve negative mitigation scores under UAP attacks, while having positive or negligible negative mitigation scores under poisoning-based attacks.

\begin{figure}[h]
    \centering
    \includegraphics[width=0.95\linewidth]{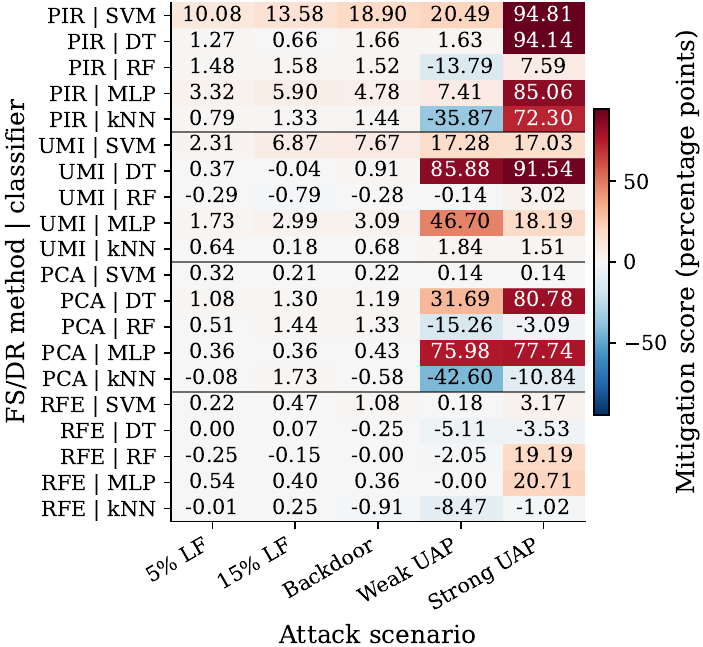}
    \caption{Overall mitigation scores based on Accuracy.}
    \label{fig:heatmap-accuracy}
\end{figure}

\begin{figure}[h]
    \centering
    \includegraphics[width=0.95\linewidth]{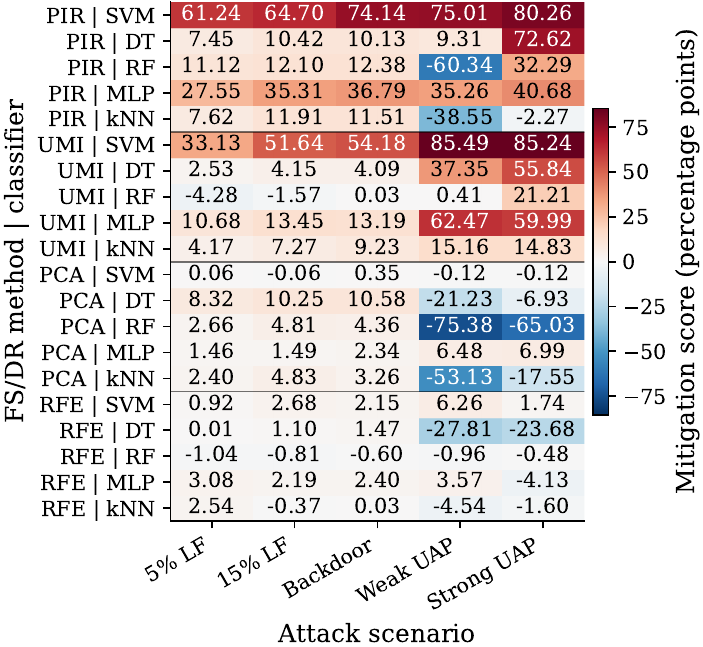}
    \caption{Overall mitigation scores based on Macro-F1.}
    \label{fig:heatmap-macro-f1}
\end{figure}

{Figures \ref{fig:heatmap-accuracy} and \ref{fig:heatmap-macro-f1} demonstrate the overall mitigation scores in terms of accuracy and Macro-F1, respectively. PIR and UMI demonstrate the most consistent robustness improvements, while PCA and RFE show more fluctuating behavior and, in many cases, reduce robustness under UAP attacks. PIR achieves positive mitigation scores in most scenarios across all the classifiers in terms of both accuracy and Macro-F1. The only cases that PIR achieves negative score are weak UAP attacks on RF and KNN for both metrics, and strong UAP attack on KNN for Macro-F1. UMI is the second most consistent method with a few more negative scores. Finally, PCA and RFE demonstrate negative mitigation scores in many scenarios, specially for Macro-F1, where the majority of scenarios demonstrate negative scores under UAP attacks.}
\section{Conclusion}
\label{sec:conclusion}
This study analyzed the performance of ML models under different adversarial attack scenarios. FS/DR techniques were then applied to compare the robustness of these models before and after feature reduction. {The results reveal that adversarial robustness is controlled by the interaction between the classifier, attack type, and feature reduction method. PIR and UMI generally provided the most consistent robustness gains across classifiers, while PCA and RFE showed more variable or negative effects in several settings. These findings demonstrate that no single FS/DR technique provides uniformly improved robustness across all the classifiers and adversarial scenarios, although some methods generally demonstrate superior performance.}

Future research will examine the effects of feature reduction techniques against other adversarial attack scenarios, such as evasion and exploratory attacks. Further investigations involving additional feature reduction techniques and more advanced ML models could provide deeper insights into their effects on adversarial robustness.
\bibliographystyle{IEEEtran}
\bibliography{main}

\end{document}